\pdfoutput=1
\documentclass[letterpaper]{article} 
\usepackage{aaai23}  
\usepackage{times}  
\usepackage{helvet}  
\usepackage{courier}  
\usepackage[hyphens]{url}  
\usepackage{graphicx} 
\usepackage{natbib}  
\usepackage{caption} 
\usepackage{algorithm}
\usepackage{algorithmic}

\usepackage{newfloat}
\usepackage{listings}
\DeclareCaptionStyle{ruled}{labelfont=normalfont,labelsep=colon,strut=off} 
\floatstyle{ruled}
\newfloat{listing}{tb}{lst}{}
\floatname{listing}{Listing}
\usepackage{subfigure}

\def\ie{\emph{i.e.}}
\def\eg{\emph{e.g.}}
\def\etal{{\em et al.\/}\, }

\usepackage{multirow}
\usepackage{amsmath,amssymb} 
\title{Text-Aware Dual Routing Network for Visual Question Answering}
\author{
    Luoqian Jiang\equalcontrib,
    Yifan He\equalcontrib,
    Jian Chen\thanks{Corresponding author.}
}
\affiliations{

    South China University of Technology\\
    \{seluoqianjiang,ellachen\}@mail.scut.edu.cn, YifanHe0112@outlook.com\\
}

\usepackage{bibentry}

\begin{document}

\maketitle

\begin{abstract}
Visual question answering (VQA) is a challenging task to provide an accurate natural language answer given an image and a natural language question about the image. It involves multi-modal learning, i.e., computer vision (CV) and natural language processing (NLP), as well as flexible answer prediction for free-form and open-ended answers. Existing approaches often fail in cases that require reading and understanding text in images to answer questions. In practice, they cannot effectively handle the answer sequence derived from text tokens because the visual features are not text-oriented. To address the above issues, we propose a Text-Aware Dual Routing Network (TDR) which simultaneously handles the VQA cases with and without understanding text information in the input images. Specifically, we build a two-branch answer prediction network that contains a specific branch for each case and further develop a dual routing scheme to dynamically determine which branch should be chosen. In the branch that involves text understanding, we incorporate the Optical Character Recognition (OCR) features into the model to help understand the text in the images. Extensive experiments on the VQA v2.0 dataset demonstrate that our proposed TDR outperforms existing methods, especially on the ``number'' related VQA questions.
\end{abstract}
\section{Introduction}

Deep neural networks have been the workhorse of many real-world applications, including image classification~\cite{guo2020breaking,guo2019nat,guo2022improving}, language modeling~\cite{dai2019deeper}, and many other areas~\cite{goodfellow2020generative,guo2020closed,guo2022towards}.
Recently, Visual Question Answering (VQA) has become an important task that combines the knowledge of computer vision and natural language processing. It has a high research value as well as a broad range of application scenarios (\eg, graphic reading, aided navigation for the blind, and medical consultation). Typically, a VQA model takes an image and a free-form, open-ended question as input and generates a reasonable answer based on visual information and natural language rules. The VQA task is extremely challenging. It involves the representation learning of different modalities (\ie, visual and linguistic modality), cross-modal fusion, and diverse answer generations. In general, there are two great challenges to be solved for the VQA task. 

First, it is non-trivial to explore the visual information from a given image. For example, many existing methods~\cite{yu2017multi,ma2018visual,hong2019exploiting} use logit (\ie, feature from the highest layer of CNN) as their visual representation, which, however, is biased towards object classification tasks, and neglects the low-level semantics such as color, texture, and number of instances. To alleviate this issue, the most recent work namely VinVL~\cite{zhang2021vinvl} obtains richer visual features by a powerful object detection model optimized with more data and annotations. Nevertheless, as VinVL considers object features only, it is hard to read and understand text in the given image, which is crucial and can not be ignored in many real-world scenes, \eg, signs and titles. In this sense, how to capture abundant visual features from images is one of the most important problems in the VQA task.

Second, how to formulate the answer prediction pattern in the VQA task still remains an open question. Specifically, to handle the answer prediction part, many recent works (\eg, \cite{zhang2021vinvl,anderson2018bottom,li2020oscar}) cast the answer prediction in VQA as a multi-label classification problem. However, with such an oversimplified problem setting, the models can only choose the answers from a predefined and fixed-length candidate set (\eg, with 3129 candidate answers), which is extremely small and can not cover all the answers in the real world. As a result, the limited number of answer candidates would further hamper the model performance as the model can only select the most likely answer in the candidate set regardless of whether this answer is true. Thus, a suitable pattern of the answer prediction is vital and necessary.

\begin{figure*}[t]
\centerline{
    \includegraphics[width=1\linewidth]{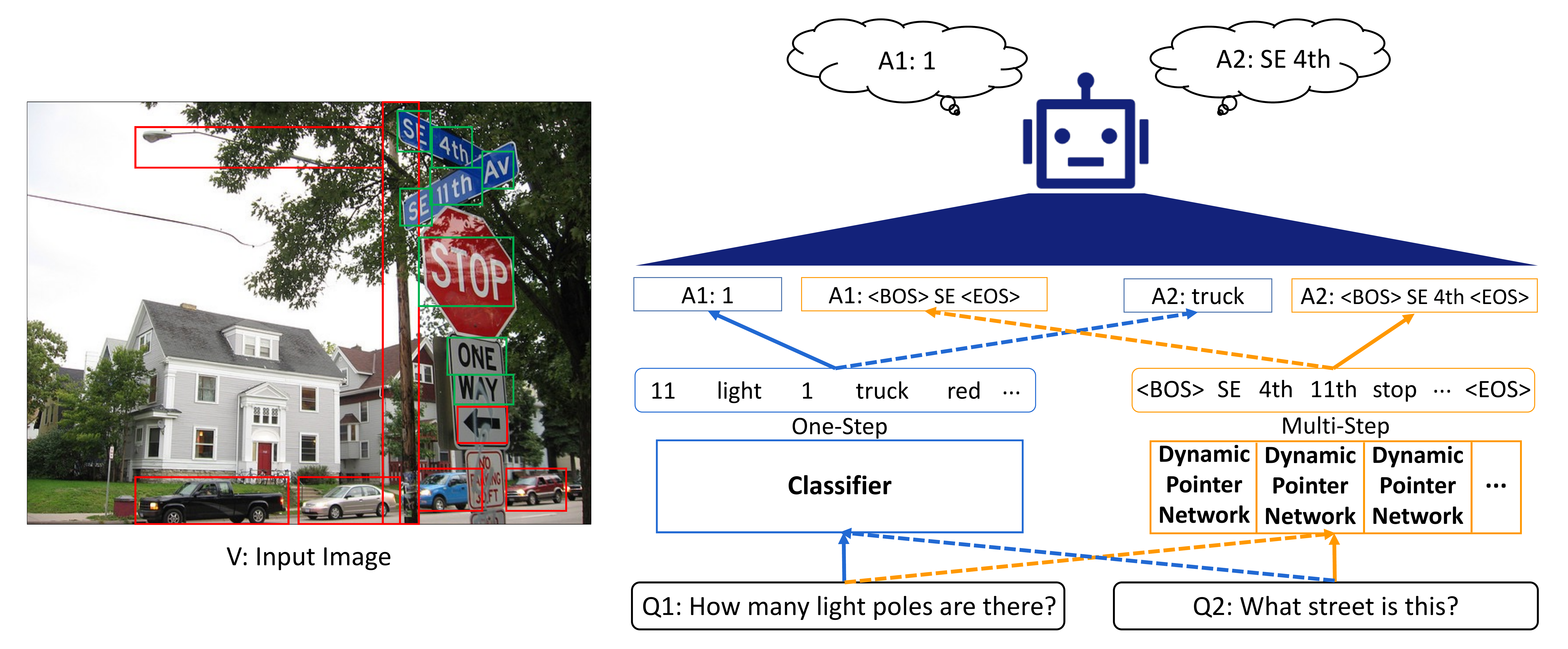}
}
    \caption{The procedure of handling the VQA task by our TDR model. Compared to the previous work, \eg, \cite{zhang2021vinvl}, our model can read the text in images. Furthermore, it employs a two-branch learning scheme and adaptively selects the final answer for different instances. Please refer to Section 3 for more details.}
    \label{TDR}
\end{figure*}

To overcome these challenges and improve performance on the VQA task, we propose a Text-Aware Dual Routing Network (TDR). We use rich object-level region features from VinVL~\cite{zhang2021vinvl} to capture appearance information from images. Because appearance information alone cannot describe the semantic information of text in images, we introduce abundant Optical Character Recognition (OCR) features as compensation. To improve visual-linguistic alignment, we organize the object tags in Oscar~\cite{li2020oscar}. The model can now observe images on a more comprehensive level thanks to the improved visual representations. To increase the flexibility in answer prediction, we propose a dual routing prediction module that incorporates a two-branch answer prediction network (\ie, a classifier and a dynamic pointer network) and a gating network. In this way, our method is not only capable of tackling general question types, but also expert in dealing with text-reading instances. In general cases, the classifier can choose a frequent answer from the candidate set. When the instance requires an answer from the text information, we propose a dynamic pointer network to copy some OCR tokens and form an answer sequence step by step during the decoding process. The gating network activates a best-match answer for each instance adaptively. Figure~\ref{TDR} depicts an intuitive procedure of our TDR model.

Our contributions are summarized as follows:
\begin{itemize}
\item We use abundant visual representations to provide a comprehensive view of images. We not only utilize richer object-level region features but also introduce OCR features as compensation on the semantic level.
\item We propose a dual routing prediction module in which we elaborately tackle text-reading and general cases with a two-branch answer prediction network. Specifically, a gating network serves as a dynamic router to activate the final answer from these two branches for each instance. 
\item Extensive experiments on the VQA v2.0 dataset demonstrate the superiority of our proposed method. Compared with the SoTA\cite{zhang2021vinvl}, our method achieves 4\% higher accuracy on ``number" answer-type instances.
\end{itemize}
\section{Related Work}

\subsection{Text-Based Visual Question Answering}
These years have seen prosperous development in the VQA task~\cite{li2020oscar,lu2019vilbert,tan2019lxmert,hu2021unit}. Text-based VQA is a sub-task of VQA that involves reading and reasoning about the text in images. To address this issue, several methods have been proposed. LoRRA~\cite{singh2019towards} attends to objects and text relevant to the question, and then chooses an answer from a fixed-length candidate set or OCR tokens with the highest attention weight. However, LoRRA predicts in a single step and cannot combine multiple tokens to form an answer sequence. To overcome this limitation, M4C~\cite{hu2020iterative} proposes a flexible pointer network to generate answers in multiple steps, with each step copied from an OCR token or a fixed vocabulary. After that, some M4C-based variants go a step further. LaAP-Net~\cite{han2020finding} uses text location as evidence to generate answers. Zhu~\etal\cite{zhu2020simple} design separate attention branches for visual and linguistic parts of text features to fuse pairwise modalities. Different from them, we use abundant visual features from object level and semantic level for image understanding. We also develop a dual routing module for answer prediction and further improve the model performance.

\subsection{Pointer Networks}
For instances involving text-reading (\eg, signs and titles) in the VQA dataset, the answers mainly come from the input OCR tokens. Pointer Networks~\cite{vinyals2015pointer} generates outputs by directly copying from the input sequence, which is not limited by a fixed-length candidate set and thus addresses the out-of-vocabulary (OOV) problem. Pointer Networks has been validated on text summarization~\cite{nallapati2016abstractive}, machine translation~\cite{gu2016incorporating}, image captioning~\cite{lu2018neural} and other tasks. Recently, Pointer Networks and its variants are applying to text-based VQA tasks. LoRRA~\cite{singh2019towards} adds dynamic OCR tokens to the candidate set for classification. M4C~\cite{hu2020iterative} incorporates the pointer network in iterative decoding, allowing each step to select an answer from the candidate set or copy a token from OCR tokens. Since the text-sensitive answers are rarely found in the candidate set, we devise a dynamic pointer network to reorganize OCR tokens into a correct answer sequence.

\subsection{Dynamic Neural Networks}
Dynamic neural networks incorporate multiple candidate modules, each of which is expert in handling different instances. A gating network is used to selectively activate these modules conditioned on the input instances~\cite{han2021dynamic}. Mixture of Experts (MoE)~\cite{shazeer2017outrageously} takes this approach by dynamically routing instances to different experts and achieves computation allocation. Lioutas~\etal\cite{lioutas2018explicit} employ a diverse group of attention experts to provide more reliable attention information than a single attention model. N. Patro~\etal\cite{patro2020deep} use a mix of experts to generate visual questions by combining multiple visual and language cues. Wang~\etal\cite{wang2021vlmo} propose to encode different input modalities by modality-specific experts. Concerning the VQA task, we find that while OCR sequence answers generally require multiple steps to generate, other answer phrases can be selected from the candidate set in a single step, implying that these two cases can be solved separately. Thus, we propose a dual routing prediction module containing two experts (\ie, a classifier and a dynamic pointer network), and then use a gating network to activate a best-match answer generated by these two experts.

\section{Method}
\subsection{Overall Framework}
\begin{figure*}[t]
\centerline{
    \includegraphics[width=\linewidth]{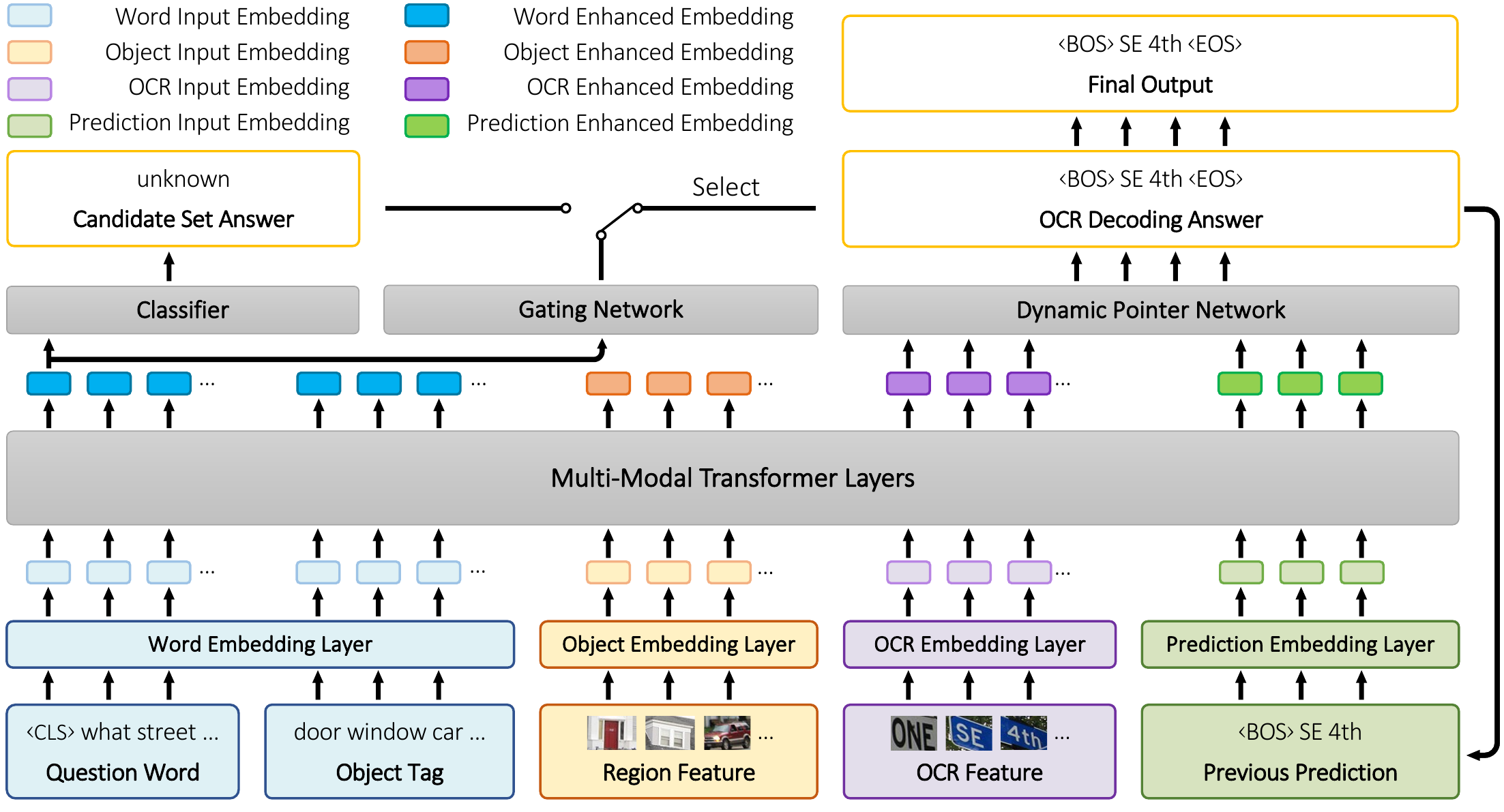}
}
    \caption{The architecture of our TDR model. In the input phase, we enhance the visual representations with abundant object features and OCR features. The input entities are all projected into a low-dimensional space before being fed into transformer layers for cross-modal fusion. Then, we propose a dual routing prediction module in which two experts are developed: a classifier for multi-label classification tasks and a dynamic pointer network to generate answers from OCR tokens via an iterative decoding process. Finally, we use a gating network to activate a best-match answer for each instance.}
    \label{fig1}
\end{figure*}
In this paper, we propose a Text-Aware Dual Routing Network (TDR) which simultaneously handles the
VQA cases with and without understanding text information in the input images. Given an image and a related question, we first improve the visual representations for image understanding by utilizing richer object-level region features and elaborate OCR features. Then, we feed all of the input features into Transformer~\cite{vaswani2017attention} for cross-modal interaction. To enhance the intelligence of our model, we propose a dual routing prediction module composed of a gating network and two experts, namely a classifier for selecting answers from the candidate set and a dynamic pointer network that generates answers from OCR tokens through an iterative decoding process. Finally, we employ a gating network to activate a best-match answer for each instance. The overall architecture of our proposed method is shown in Figure~\ref{fig1}.

\subsection{Visual Representation}

\subsubsection{Object Feature}

Visual representations are essential for the advancement of VQA tasks. However, many methods rely on region features proposed in~\cite{anderson2018bottom} to identify salient regions from detected objects without further optimizing the visual features. Recently, VinVL~\cite{zhang2021vinvl} has trained a new object detector on a large-scale corpus to extract richer visual features, which significantly improves SoTA results. Given an instance (\ie, an image and a related question), we use the object-level region features extracted by VinVL and project them into a series of object embeddings $\{h^{obj}_e\} \in \mathbb{R}^d$, where $e = 1, ..., E$. Furthermore, to enhance visual-linguistic alignment according to~\cite{li2020oscar}, we append the detected object tags to the question words and embed them in $d$-dimensional word embeddings as $\{h^{word}_v\} \in \mathbb{R}^d$, where $v = 1, ..., V$. 


\subsubsection{OCR Feature}
OCR tokens extracted from images necessarily require special consideration, as they contain visual and semantic information critical to understanding the text in images and cannot be expressed solely through region features. Given an image with $M$ OCR tokens derived from the Rosetta system~\cite{borisyuk2018rosetta}, we introduce the OCR feature for each OCR patch following the setting in~\cite{hu2020iterative}. The OCR feature consists of two parts (\ie, semantic and spatial ones), which focus on the content and position of each OCR patch, respectively. Regarding the $m$-th OCR token (where $m = 1, ..., M$), to better capture the semantic information from the OCR patch, we further divide the semantic part ${x}^{smt}_m$ into a FastText feature~\cite{bojanowski2017enriching}, a Pyramidal Histogram of Characters (PHOC) feature~\cite{almazan2014word}, and an appearance feature. Specifically, the FastText feature ${x}^{ft}_m \in \mathbb{R}^{300}$ constructs word embedding using subword information, which endows the model with the ability to handle rare and unknown words. The PHOC feature ${x}^{p}_m \in \mathbb{R}^{604}$ is a character-level representation while the appearance feature ${x}^{fr}_m \in \mathbb{R}^{2048}$ is an object-level representation detected by Faster R-CNN~\cite{ren2015faster}, which primarily reflects the color, font, and background information of the OCR patch. Formally, to obtain the semantic part ${x}^{smt}_m$, we compose these features by:
\begin{equation}
{x}^{smt}_m= [{x}^{ft}_m ; {x}^p_m ; {x}^{fr}_m],
\end{equation}
where $[\cdot;\cdot;\cdot]$ denotes concatenation operation. On the other hand, the spatial part ${x}^{spt}_m \in \mathbb{R}^{4}$ refers to the relative position of the OCR patch in the given image, which is denoted as bounding box coordinates. After that, we feed both the ${x}^{smt}_m$ and ${x}^{spt}_m$ features into an OCR embedding layer and obtain the $d$-dimensional integrated OCR embeddings $\{{h}^{ocr}_m\} \in \mathbb{R}^d$ (where $m = 1, ..., M$) as follows:
\begin{equation}
{h}^{ocr}_m=LN({W}^{smt}{x}^{smt}_m)+LN({W}^{spt}{x}^{spt}_m),
\end{equation}
where ${W}^{smt}$ and ${W}^{spt}$ are learnable weights while $LN(\cdot)$ denotes layer normalization.

\subsection{Cross-Modal Interaction}
The multi-modal transformer layers are built with an encoder-decoder architecture, and the multi-head attention mechanism allows for inter-modal and intra-modal interaction. Entities in each input modality can attend to one another while each decoding step can attend to all positions preceding the current step. We feed a sequence of $d$-dimensional inputs into the transformer layers, namely word embeddings $\{h^{word}_v\}$, object embeddings $\{h^{obj}_e\}$, OCR embeddings $\{h^{ocr}_m\}$, and previous prediction embeddings $\{h^{dec}_t\}$ (where the decoding step $t = 1, ..., T$) to produce a set of $d$-dimensional enhanced representations. Specifically, $[CLS]$ is a special token added to the beginning of each input sequence, and ${z}^{[CLS]}$ is used as the aggregated representation for classification tasks. In the case of sequence-to-sequence learning, the decoder generates a sequence of answers based on the auto-regressive scheme. At each time step, we feed some enhanced representations and a previous prediction embedding into a dynamic pointer network to predict the next step (explained in the next section).

\subsection{Dual Routing Prediction Module}
The VQA dataset contains open-ended questions with flexible answers. In practice, there are two types of questions: general questions that can be answered with a common answer set and questions that must be answered using the text information in the image. To improve the intelligence of our model, we propose a two-branch answer prediction network. Since the first case can be thought of as a multi-label classification problem, we use a classifier to select an entire answer (single-word or multi-word) in one step. In the second case, multi-step decoding is required to generate a sequence of answers. To this end, we propose a dynamic pointer network to copy an answer token from OCR tokens in each step. Finally, a gating network acts as a router to activate a best-match answer generated by these two experts for each instance.

\subsubsection{Two-Branch Answer Prediction Network}

\paragraph{Classifier}
According to~\cite{anderson2018bottom}, in general cases, we can cast the answer prediction in VQA as a multi-label classification problem and use a set of $C$ frequent answers in the VQA v2.0 training set as the candidate set. Assume we have $N$ instances in the training set, for the $i$-th (where $i = 1, ..., N$) instance, we perform a linear transformation on the aggregated representation ${z}_i^{[CLS]}$ to calculate the predicted score $\hat{s}_{i}$ for these candidate answers. To train the classifier, we compute the Binary Cross Entropy (BCE) loss $\mathcal{L}_{cls}$ between the predicted and ground-truth scores of the instance taking this branch as follows:
\begin{equation}
   \mathcal{L}_{cls}=-\frac{1}{N}\sum^{N}_{i=1}g_i\sum^{C}_{c=1}[s_{i,c}log(\hat{s}_{i,c})+(1-s_{i,c})log(1-\hat{s}_{i,c})],
\label{clsloss}
\end{equation}
where the ground-truth routing flag $g_i=1$ indicates that the instance selects the answer generated by the classifier, otherwise $g_i=0$. $C$ is the fixed length of the candidate set, and $s_{i,c}$ is the ground-truth answer score calculated by Eq.~(\ref{metric}) below. During inference, we select the answer index with the highest predicted score as $argmax\ \hat{\mathbf{s}}_i$.

In this way, we can solve common cases where the answer is not text-sensitive (\eg, object detection and counting) with low complexity. However, when the instance asks for the text in the image, the capacity of OCR tokens in diverse images is too large for a fixed-length candidate set. The classifier cannot combine these tokens to generate a specific answer sequence. To address these issues, another prediction approach should be proposed in addition to the classifier.


\paragraph{Dynamic Pointer Network}
Some answers in the VQA v2.0 dataset are derived directly from image text and composed of OCR tokens (\eg, signs and titles). Because it is impossible to include all of these diverse OCR tokens in the candidate set, we consider dynamically selecting and combining OCR tokens for each image to generate a sequence of answers. According to pointer networks~\cite{vinyals2015pointer}, one can copy tokens directly from the input text, which simplifies the answer generation process and avoids OOV problems. During the $T$-step decoding process, each previous prediction embedding ${h}^{dec}_t$ ($t = 1, ..., T$) is the sum of the token embedding, the positional embedding, and the type embedding. We feed them into the transformer layers to get a set of enhanced representations $\{{z}_t^{dec}\} \in \mathbb{R}^d$ (where $t = 1, ..., T$). For the $t$-th step, we calculate the score $y^t_{m}$ for the $m$-th (where $m = 1, ..., M$) OCR token using a bilinear interaction between ${z}_t^{dec}$ and ${z}_m^{ocr}$ as follows:
\begin{equation}
    y^t_{m}=({W}^{ocr}{z}_m^{ocr}+{b}^{ocr})^T({W}^{dec}{z}_t^{dec}+{b}^{dec}),
\label{ocrscore}
\end{equation}
where ${W}^{ocr}$ and ${W}^{dec} \in \mathbb{R}^{d \times d}$ denote weight matrices, ${b}^{ocr}$ and ${b}^{dec} \in \mathbb{R}^{d}$ denote bias vectors. In the $t$-th step, we select an answer index with the highest predicted score as $argmax\ \{y_1^t, y_2^t, ..., y_M^t\}$, and select an answer word from these M OCR tokens with it.

Since instances that cannot find answers in the candidate set or OCR tokens are eliminated from the training set, the remaining $\sum^{N}_{i=1}(1-g_i)$ instances follow the pointer network branch. For each step of the decoding process, we calculate the BCE loss $\mathcal{L}_{ptr}$ between the predicted and ground-truth scores. Specifically, the score for the first ground-truth token is calculated by Eq.~(\ref{metric}), and each subsequent token is filled with a full score of 1.0.
\begin{gather}
   {loss}_{i} =s^t_{i,m}log(y^t_{i,m})+(1-s^t_{i,m})log(1-y^t_{i,m}) \notag \\
  \mathcal{L}_{ptr}=-avg_{ptr}(\sum^{N}_{i=1}(1-g_i)\sum^{T}_{t=1}\sum^{M}_{m=1}loss_i),
\label{ptrloss}
\end{gather}
where $N$ is the number of training set instances, $T$ is the maximum decoding length, and $M$ is the maximum capacity of OCR tokens detected in an image. $s^t_{i,m}$ is the score of the ground-truth answer token for the $t$-th step, and $y^t_{i,m}$ is the predicted OCR score in Eq.~(\ref{ocrscore}). Because the valid decoding length for each instance may differ, $avg_{ptr}(\cdot)$ averages the loss from all valid decoding steps of these instances.

\subsubsection{Gating Network}
The questions in the VQA task are open-ended, requiring the model to use a variety of abilities and knowledge to solve them. We divide the instances of the dataset into two categories based on whether they are answered with the text in images. The aforementioned dynamic pointer network handles cases where an answer sequence must be generated using detected OCR tokens in the image, while the classifier handles other common cases. To deal with this dynamic routing process, a gating network is proposed. The gating network is essentially a Multi-Layer Perception network (MLP) designed for binary classification. For the $i$-th (where $i = 1, ..., N$) instance, we take ${z}_i^{[CLS]}$ as input and compute the gating score $\hat{g}_i$ as follows:
\begin{equation}
  \hat{g}_i=\sigma(W_2ReLU(W_1z_i^{[CLS]}+b_1)+b_2),
\end{equation}
where $W_i$ and $b_i$ are learnable parameters, and $\sigma(\cdot)$ is the sigmoid function that projects the value into the range [0, 1]. During inference, we use the ceiling function to convert $\hat{g}_i$ to discrete values 0 and 1 as a routing indicator to activate a best-match answer from the above two branches, \ie, the classifier and the dynamic pointer network. The gating network is trained with the BCE loss $\mathcal{L}_{gate}$ as follows:
\begin{equation}
   \mathcal{L}_{gate}=-\frac{1}{N}\sum^N_{i=1}[g_ilog(\hat{g}_i)+(1-g_i)log(1-\hat{g}_i)],
\label{gateloss}
\end{equation}
where $N$ is the number of training set instances, and $g_i$ denotes the ground-truth routing flag with a value of 0 or 1.

\begin{figure*}[t]
\begin{minipage}[t]{0.5\columnwidth}
\includegraphics[width=\linewidth]{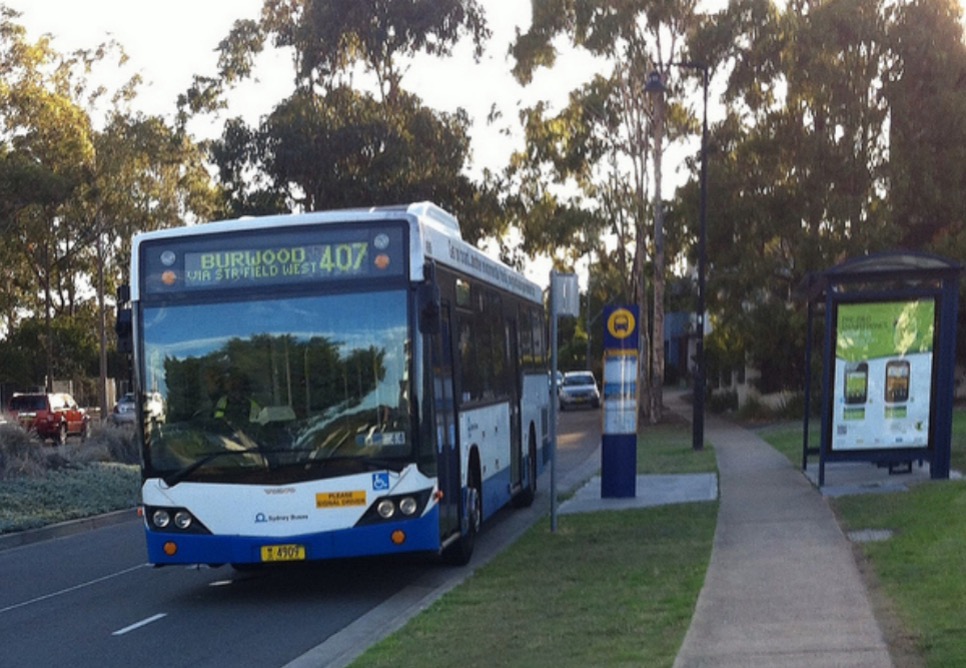}
 Q1: What number is lit up on the bus?\\\\
 {\textbf{VinVL}:} 1\\
 {\textbf{TDR (ours)}:} 407\\
 \textbf{Human}: 407
\end{minipage}
\begin{minipage}[t]{0.5\columnwidth}
\includegraphics[width=\linewidth]{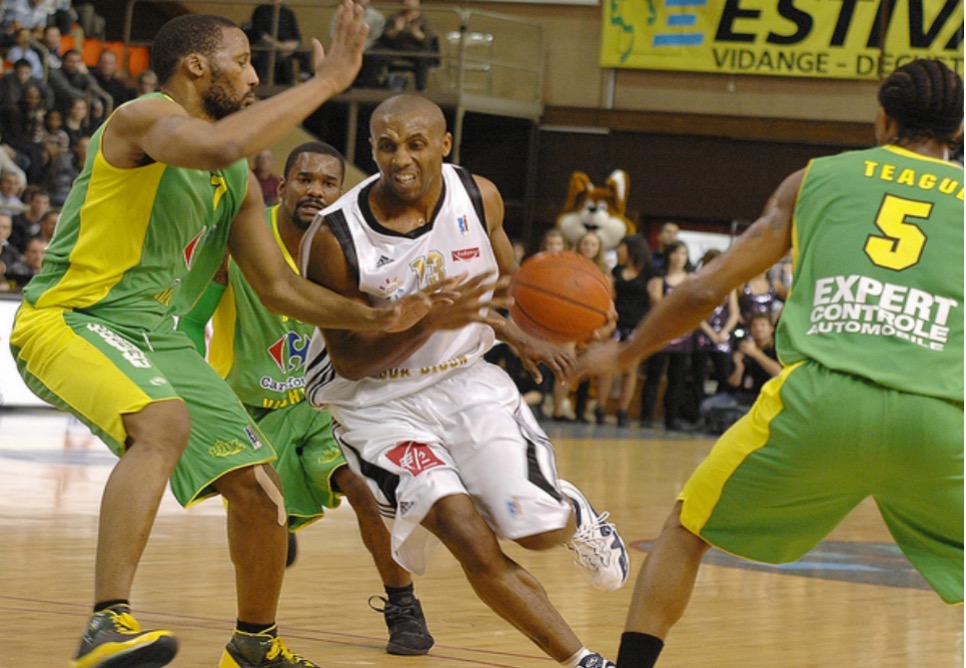}
        Q2: What number is on the right man's shirt?\\\\
        {\textbf{VinVL}:} 7\\
        {\textbf{TDR (ours)}:} 5\\
        \textbf{Human}: 5
\end{minipage}
\begin{minipage}[t]{0.5\columnwidth}
\includegraphics[width=\linewidth]{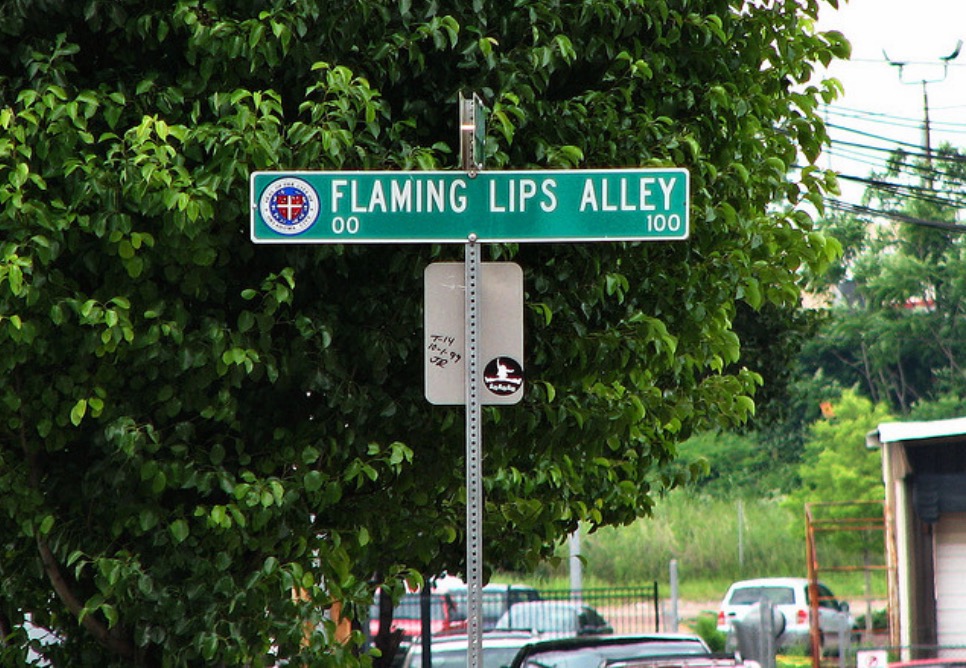}
        Q3: What is the name of this street?\\\\
        {\textbf{VinVL}:} main\\
        {\textbf{TDR (ours)}:} FLAMING\\LIPS ALLEY\\
        \textbf{Human}: FLAMING\\LIPS ALLEY
\end{minipage}
\begin{minipage}[t]{0.5\columnwidth}
\includegraphics[width=\linewidth]{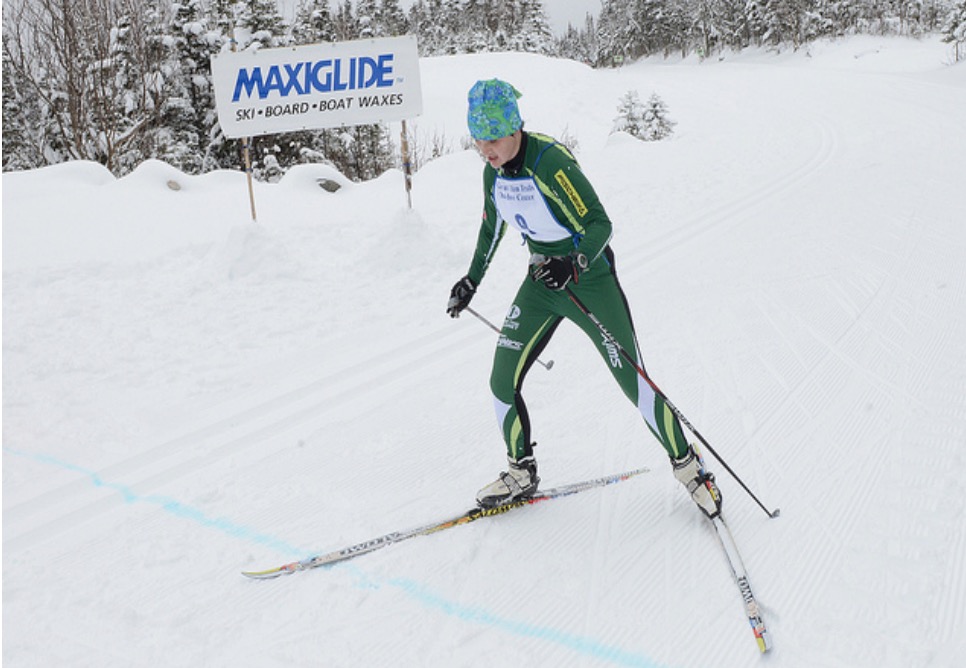}
        Q4: What is written on the flag?\\\\
        {\textbf{VinVL}:} nothing\\
        {\textbf{TDR (ours)}:} MAXIGLIDE\\
        \textbf{Human}: MAXIGLIDE
\end{minipage}
\caption{Qualitative examples on the VQA v2.0 validation set.}
\label{fig3}
\end{figure*} 
\subsection{Overall Objective Function}
To eliminate the impact of imbalance distribution in the proposed two branches, we use the weighted version for $\mathcal{L}_{cls}$ in Eq.~(\ref{clsloss}) and $\mathcal{L}_{ptr}$ in Eq.~(\ref{ptrloss}). Finally, the objective function is formatted as follows:
\begin{equation}
    \mathcal{L}=\omega_{cls}\mathcal{L}_{cls}+\omega_{ptr}\mathcal{L}_{ptr}+\mathcal{L}_{gate},
\label{allloss}
\end{equation}
where $\omega_{cls}=\frac{\sum^{N}_{i=1}g_i}{N}$, $\omega_{ptr}=\frac{\sum^{N}_{i=1}(1-g_i)}{N}$.

\section{Experiments}
\subsection{Dataset}
The VQA v2.0 dataset~\cite{goyal2017making} includes 265016 images from COCO and abstract scenes, at least 3 open-ended questions per image, and 10 ground-truth answers per question. It is a well-balanced dataset that provides two similar images with distinct answers to the same question. These instances require a detailed visual understanding of images to extract crucial information for reasoning. As a comprehensive dataset, VQA v2.0 covers a wide range of question types, and about 43\% of the images contain essential text information. Therefore, a VQA model should apply a variety of skills to complete the task intelligently. 
\subsection{Evaluation Metric}
We use the evaluation metric proposed in~\cite{antol2015vqa} that is robust to inter-human variable answers as follows: 
\begin{equation}
    Acc(ans)=min\{\frac{\# humans\ that\ said\ ans}{3}, 1\}.
    \label{metric}
\end{equation}
As shown in Eq.~(\ref{metric}), if at least three humans give the same answer, the answer is considered 100\% accurate. Before evaluation, a series of regularization operations are performed, such as converting all answers to lowercase, using Arabic numerals uniformly, and removing punctuation marks and articles.

\subsection{Implementation Details}
Our models are implemented in PyTorch and optimized using the Adam optimizer. We fine-tune our BERT-base and BERT-large models with initial learning rates of 5e-05 and 1e-05, respectively. The maximum length of tokenized words is $V$ = 128. We detect at most $E$ = 50 objects and $M$ = 50 OCR tokens per image. The maximum decoding step is $T$ = 12. The dimensionality of the embedding space is set to $d$ = 768. The candidate set has a size of $C$ = 3129. The dropout rate is 0.3 and the weight decay is 0.05. We use 12 attention heads in transformer layers and leave the other hyper-parameters unchanged from BERT~\cite{kenton2019bert}. Due to hardware limitations, we run the training procedure for 35 epochs with a batch size of 48. 
\subsection{Comparison with State-of-The-Art Methods}
We compare TDR with previous state-of-the-art methods including UNITER~\cite{chen2019uniter}, VILLA~\cite{gan2020large}, ERNIE-VIL~\cite{yu2021ernie}, Oscar~\cite{li2020oscar} and the strongest baseline VinVL~\cite{zhang2021vinvl} on the VQA v2.0 testing set (\ie, ``test-dev" and ``test-std"). Table~\ref{table1} gives an overall comparison of BERT-base size and BERT-large size models. TDR outperforms all the other methods and especially surpasses VinVL by around 0.8\% under the same experimental setting. To further validate the effect on different answer-type instances, we compare TDR with VinVL in Table~\ref{table2}, we find that our final model (line 3) leads to over 4\% improvement on ``number" answer-type instances of VQA v2.0 dataset (line 1).


\begin{table}[htbp]
  \centering
    \begin{tabular}{c|cc|cc}
    \hline
     \multirow{2}{*}{Dataset} & \multicolumn{4}{c}{VQA v2.0} \\
     \cline{2-5}
     &\multicolumn{2}{c|}{Test-dev} & 
     \multicolumn{2}{c}{Test-std} \\
    \hline
    Model Size & Base  & Large & Base  & Large \\
    \hline
    UNITER & 72.27 & 73.24 & 72.46 & 73.40 \\
    \hline
    VILLA & 73.59 & 73.69 & 73.67 & 74.87 \\
    \hline
    ERNIE-VIL & 72.62 & 74.75 & 72.85 & 74.93 \\
    \hline
    Oscar & 73.16 & 73.61 & 73.44 & 73.82 \\
    \hline
    VinVL & 74.78 & 76.04 & 74.87 & 76.06 \\
    \hline
    TDR (ours) & \textbf{75.56} & \textbf{76.73} & \textbf{75.63} &\textbf{76.87} \\
    \hline
    \end{tabular}%
  \caption{Performance comparison on the VQA v2.0 dataset. (``Base" and ``Large" indicate models of BERT-base and BERT-large size.)}
  \label{table1}%
\end{table}%
\begin{table*}[t]
  \centering
    \begin{tabular}{c|c|c|cc|cc|cc|cc}
    \hline
    \multirow{2}{*}{Method} & {OCR} & {Dual} & \multicolumn{2}{c|}{Yes/No} & \multicolumn{2}{c|}{Number} & \multicolumn{2}{c|}{Other} & \multicolumn{2}{c}{All} \\

         &Feature       &Routing       & Dev   & Std   & Dev   & Std   & Dev   & Std   & Dev   & Std \\
    \hline
        {VinVL} &   ---   &   ---    & {90.62} & {90.73} & {56.28} & {55.63} & 65.45 & 65.47 & 74.78 & {74.87} \\
    \hline
    \multirow{2}{*}{TDR} &{\checkmark}     &       & {\textbf{90.92}} & {90.84} & {57.58}& {57.57} & {65.44} & {65.69} & {75.04} & {75.23} \\
{} &{\checkmark}     &{\checkmark}       & {90.64} & {\textbf{90.92}} & {\textbf{60.37}} &{\textbf{59.29}} & {\textbf{66.11}} & {\textbf{66.06}} & {\textbf{75.56}} & {\textbf{75.63}} \\

    \hline
    \end{tabular}%
  \caption{Performance on different answer-type instances of the VQA v2.0 dataset. (Test-dev is denoted by ``Dev" and Test-std is denoted by ``Std".)}
  \label{table2}%
\end{table*}%

\subsection{Effect of OCR Feature}
To investigate the effect of the OCR feature, we conduct an ablation experiment with models of the BERT-base size in Table~\ref{table2}. Unlike VinVL, the restricted version of our TDR model (line 2) incorporates OCR features in the visual representation. In the answer prediction phase, both models use the classifier to predict an entire answer in one step. Comparing line 2 to 1, OCR features lead to an overall improvement, particularly in the ``number" answer type known to be difficult for VQA models~\cite{lu2019vilbert}. We especially uplift the performance of text-reading instances (\eg, ``What number is ... ?", ``What is ... ?"), demonstrating the validity of incorporating OCR features as supplement information.
\subsection{Effect of Dual Routing Prediction Module}
As shown in Table~\ref{table2}, we train our full TDR model with a dual routing prediction module (line 3), which gives another 2\% improvement on ``number" answer-type instances compared to its counterpart using a classifier for answer prediction (line 2). When compared to VinVL, our full TDR model achieves around 4\% improvement (line 3 VS line 1) on this answer type. 

The dual routing prediction module incorporates a two-branch answer prediction network (\ie, a classifier and a dynamic pointer network) and a gating network. To intuitively show the effect of the two-branch learning scheme on tackling text-reading and general cases, we evaluate our TDR and VinVL on different answer-source instances in Table~\ref{table3}. Specifically, we divide the validation set into two categories based on the answer source, \ie, the final answer is either from the OCR tokens or from the candidate set, and then calculate the average score using Eq.~(\ref{metric}). Compared with VinVL, our model can generate much more accurate OCR sequence answers in text-reading cases, improving the performance by a significant margin (\textgreater 30\%), while ensuring the accuracy of selecting an answer from the candidate set.
\begin{table}[htbp]
  \centering
    \begin{tabular}{c|c|cr}
\hline    Answer Source & OCR Token & Candidate Set\\
\hline   {VinVL} & 22.08 & 76.28 \\
\hline   {TDR (ours)} & 58.23 & 75.69 \\
\hline    \end{tabular}%
  \caption{Performance on different answer-source instances of the VQA v2.0 dataset.}
  \label{table3}%
\end{table}%

Essentially, the gating network performs a binary classification task to finally determine the final answer from the two branches mentioned above. To assess the performance of the gating network, we use commonly used classification task metrics, \ie, accuracy and F1-score. A high value of these metrics indicates good performance. As a result, we get 98.7\% accuracy and an F1-score of 0.81. 

\subsection{Qualitative Analysis}
The major failure cases of existing methods in the VQA v2.0 dataset involve text reading and answering. As shown in Figure~\ref{fig3}, the SoTA VinVL degrades when accounting questions such as ``What number is ... ?" and ``What is ... ?". Under these difficult conditions, our TDR model can still correctly identify and reorganize OCR tokens in the image to generate a complete answer sequence, proving the advancement of our proposed method.


\section{Conclusion} 
In this paper, we propose a Text-Aware Dual Routing Network (TDR) for the VQA task. We enrich the visual representations with abundant region features and OCR features to provide a comprehensive visual-linguistic view of the input image. Thus, the model can utilize text information as supplementary knowledge for reasoning. Furthermore, we propose a dual routing prediction module incorporating a two-branch answer prediction network and a gating network to improve the flexibility of answer prediction. In this way, we can simultaneously handle the VQA cases with and without understanding text information in the input images. We not only improve the diversity and the accuracy of the generated answers but also inspire a new way of thinking about achieving human intelligence. Extensive experiments on the VQA v2.0 dataset demonstrate the superiority of our method over the considered baseline methods. We significantly improve the performance on the ``number” related questions known to be difficult for existing VQA models.

\bibliography{aaai23}

\begin{thebibliography}{40}
\providecommand{\natexlab}[1]{#1}

\bibitem[{Almaz{\'a}n et~al.(2014)Almaz{\'a}n, Gordo, Forn{\'e}s, and
  Valveny}]{almazan2014word}
Almaz{\'a}n, J.; Gordo, A.; Forn{\'e}s, A.; and Valveny, E. 2014.
\newblock Word spotting and recognition with embedded attributes.
\newblock \emph{IEEE transactions on pattern analysis and machine
  intelligence}, 36(12): 2552--2566.

\bibitem[{Anderson et~al.(2018)Anderson, He, Buehler, Teney, Johnson, Gould,
  and Zhang}]{anderson2018bottom}
Anderson, P.; He, X.; Buehler, C.; Teney, D.; Johnson, M.; Gould, S.; and
  Zhang, L. 2018.
\newblock Bottom-up and top-down attention for image captioning and visual
  question answering.
\newblock In \emph{Proceedings of the IEEE conference on computer vision and
  pattern recognition}, 6077--6086.

\bibitem[{Antol et~al.(2015)Antol, Agrawal, Lu, Mitchell, Batra, Zitnick, and
  Parikh}]{antol2015vqa}
Antol, S.; Agrawal, A.; Lu, J.; Mitchell, M.; Batra, D.; Zitnick, C.~L.; and
  Parikh, D. 2015.
\newblock Vqa: Visual question answering.
\newblock In \emph{Proceedings of the IEEE international conference on computer
  vision}, 2425--2433.

\bibitem[{Bojanowski et~al.(2017)Bojanowski, Grave, Joulin, and
  Mikolov}]{bojanowski2017enriching}
Bojanowski, P.; Grave, E.; Joulin, A.; and Mikolov, T. 2017.
\newblock Enriching word vectors with subword information.
\newblock \emph{Transactions of the association for computational linguistics},
  5.

\bibitem[{Borisyuk, Gordo, and Sivakumar(2018)}]{borisyuk2018rosetta}
Borisyuk, F.; Gordo, A.; and Sivakumar, V. 2018.
\newblock Rosetta: Large scale system for text detection and recognition in
  images.
\newblock In \emph{Proceedings of the 24th ACM SIGKDD international conference
  on knowledge discovery \& data mining}.

\bibitem[{Chen et~al.(2019)Chen, Li, Yu, El~Kholy, Ahmed, Gan, Cheng, and
  Liu}]{chen2019uniter}
Chen, Y.-C.; Li, L.; Yu, L.; El~Kholy, A.; Ahmed, F.; Gan, Z.; Cheng, Y.; and
  Liu, J. 2019.
\newblock Uniter: Learning universal image-text representations.

\bibitem[{Dai and Callan(2019)}]{dai2019deeper}
Dai, Z.; and Callan, J. 2019.
\newblock Deeper text understanding for IR with contextual neural language
  modeling.
\newblock In \emph{Proceedings of the 42nd International ACM SIGIR Conference
  on Research and Development in Information Retrieval}.

\bibitem[{Gan et~al.(2020)Gan, Chen, Li, Zhu, Cheng, and Liu}]{gan2020large}
Gan, Z.; Chen, Y.-C.; Li, L.; Zhu, C.; Cheng, Y.; and Liu, J. 2020.
\newblock Large-scale adversarial training for vision-and-language
  representation learning.
\newblock \emph{Advances in Neural Information Processing Systems}, 33:
  6616--6628.

\bibitem[{Goodfellow et~al.(2020)Goodfellow, Pouget-Abadie, Mirza, Xu,
  Warde-Farley, Ozair, Courville, and Bengio}]{goodfellow2020generative}
Goodfellow, I.; Pouget-Abadie, J.; Mirza, M.; Xu, B.; Warde-Farley, D.; Ozair,
  S.; Courville, A.; and Bengio, Y. 2020.
\newblock Generative adversarial networks.
\newblock \emph{Communications of the ACM}, 63(11): 139--144.

\bibitem[{Goyal et~al.(2017)Goyal, Khot, Summers-Stay, Batra, and
  Parikh}]{goyal2017making}
Goyal, Y.; Khot, T.; Summers-Stay, D.; Batra, D.; and Parikh, D. 2017.
\newblock Making the v in vqa matter: Elevating the role of image understanding
  in visual question answering.
\newblock In \emph{Proceedings of the IEEE conference on computer vision and
  pattern recognition}, 6904--6913.

\bibitem[{Gu et~al.(2016)Gu, Lu, Li, and Li}]{gu2016incorporating}
Gu, J.; Lu, Z.; Li, H.; and Li, V.~O. 2016.
\newblock Incorporating copying mechanism in sequence-to-sequence learning.
\newblock \emph{arXiv preprint arXiv:1603.06393}.

\bibitem[{Guo et~al.(2020{\natexlab{a}})Guo, Chen, Wang, Chen, Cao, Deng, Xu,
  and Tan}]{guo2020closed}
Guo, Y.; Chen, J.; Wang, J.; Chen, Q.; Cao, J.; Deng, Z.; Xu, Y.; and Tan, M.
  2020{\natexlab{a}}.
\newblock Closed-loop matters: Dual regression networks for single image
  super-resolution.
\newblock In \emph{IEEE/CVF conference on computer vision and pattern
  recognition}.

\bibitem[{Guo et~al.(2020{\natexlab{b}})Guo, Chen, Zheng, Zhao, Chen, Huang,
  and Tan}]{guo2020breaking}
Guo, Y.; Chen, Y.; Zheng, Y.; Zhao, P.; Chen, J.; Huang, J.; and Tan, M.
  2020{\natexlab{b}}.
\newblock Breaking the curse of space explosion: Towards efficient nas with
  curriculum search.
\newblock In \emph{International Conference on Machine Learning}. PMLR.

\bibitem[{Guo, Stutz, and Schiele(2022)}]{guo2022improving}
Guo, Y.; Stutz, D.; and Schiele, B. 2022.
\newblock Improving robustness by enhancing weak subnets.
\newblock In \emph{European Conference on Computer Vision}. Springer.

\bibitem[{Guo et~al.(2022)Guo, Wang, Chen, Cao, Deng, Xu, Chen, and
  Tan}]{guo2022towards}
Guo, Y.; Wang, J.; Chen, Q.; Cao, J.; Deng, Z.; Xu, Y.; Chen, J.; and Tan, M.
  2022.
\newblock Towards lightweight super-resolution with dual regression learning.
\newblock \emph{arXiv preprint arXiv:2207.07929}.

\bibitem[{Guo et~al.(2019)Guo, Zheng, Tan, Chen, Chen, Zhao, and
  Huang}]{guo2019nat}
Guo, Y.; Zheng, Y.; Tan, M.; Chen, Q.; Chen, J.; Zhao, P.; and Huang, J. 2019.
\newblock Nat: Neural architecture transformer for accurate and compact
  architectures.
\newblock \emph{Advances in Neural Information Processing Systems}, 32.

\bibitem[{Han, Huang, and Han(2020)}]{han2020finding}
Han, W.; Huang, H.; and Han, T. 2020.
\newblock Finding the evidence: Localization-aware answer prediction for text
  visual question answering.
\newblock \emph{arXiv preprint arXiv:2010.02582}.

\bibitem[{Han et~al.(2021)Han, Huang, Song, Yang, Wang, and
  Wang}]{han2021dynamic}
Han, Y.; Huang, G.; Song, S.; Yang, L.; Wang, H.; and Wang, Y. 2021.
\newblock Dynamic neural networks: A survey.
\newblock \emph{IEEE Transactions on Pattern Analysis and Machine
  Intelligence}.

\bibitem[{Hong et~al.(2019)Hong, Fu, Uh, Mei, and Byun}]{hong2019exploiting}
Hong, J.; Fu, J.; Uh, Y.; Mei, T.; and Byun, H. 2019.
\newblock Exploiting hierarchical visual features for visual question
  answering.
\newblock \emph{Neurocomputing}, 351: 187--195.

\bibitem[{Hu and Singh(2021)}]{hu2021unit}
Hu, R.; and Singh, A. 2021.
\newblock Unit: Multimodal multitask learning with a unified transformer.
\newblock In \emph{Proceedings of the IEEE/CVF International Conference on
  Computer Vision}.

\bibitem[{Hu et~al.(2020)Hu, Singh, Darrell, and Rohrbach}]{hu2020iterative}
Hu, R.; Singh, A.; Darrell, T.; and Rohrbach, M. 2020.
\newblock Iterative answer prediction with pointer-augmented multimodal
  transformers for textvqa.
\newblock In \emph{Proceedings of the IEEE/CVF Conference on Computer Vision
  and Pattern Recognition}.

\bibitem[{Kenton and Toutanova(2019)}]{kenton2019bert}
Kenton, J. D. M.-W.~C.; and Toutanova, L.~K. 2019.
\newblock BERT: Pre-training of Deep Bidirectional Transformers for Language
  Understanding.
\newblock In \emph{Proceedings of NAACL-HLT}.

\bibitem[{Li et~al.(2020)Li, Yin, Li, Zhang, Hu, Zhang, Wang, Hu, Dong, Wei
  et~al.}]{li2020oscar}
Li, X.; Yin, X.; Li, C.; Zhang, P.; Hu, X.; Zhang, L.; Wang, L.; Hu, H.; Dong,
  L.; Wei, F.; et~al. 2020.
\newblock Oscar: Object-semantics aligned pre-training for vision-language
  tasks.
\newblock In \emph{European Conference on Computer Vision}. Springer.

\bibitem[{Lioutas, Passalis, and Tefas(2018)}]{lioutas2018explicit}
Lioutas, V.; Passalis, N.; and Tefas, A. 2018.
\newblock Explicit ensemble attention learning for improving visual question
  answering.
\newblock \emph{Pattern Recognition Letters}, 111: 51--57.

\bibitem[{Lu et~al.(2019)Lu, Batra, Parikh, and Lee}]{lu2019vilbert}
Lu, J.; Batra, D.; Parikh, D.; and Lee, S. 2019.
\newblock Vilbert: Pretraining task-agnostic visiolinguistic representations
  for vision-and-language tasks.
\newblock \emph{Advances in neural information processing systems}, 32.

\bibitem[{Lu et~al.(2018)Lu, Yang, Batra, and Parikh}]{lu2018neural}
Lu, J.; Yang, J.; Batra, D.; and Parikh, D. 2018.
\newblock Neural baby talk.
\newblock In \emph{Proceedings of the IEEE conference on computer vision and
  pattern recognition}, 7219--7228.

\bibitem[{Ma et~al.(2018)Ma, Shen, Dick, Wu, Wang, van~den Hengel, and
  Reid}]{ma2018visual}
Ma, C.; Shen, C.; Dick, A.; Wu, Q.; Wang, P.; van~den Hengel, A.; and Reid, I.
  2018.
\newblock Visual question answering with memory-augmented networks.
\newblock In \emph{IEEE conference on computer vision and pattern recognition}.

\bibitem[{Nallapati et~al.(2016)Nallapati, Zhou, Gulcehre, Xiang
  et~al.}]{nallapati2016abstractive}
Nallapati, R.; Zhou, B.; Gulcehre, C.; Xiang, B.; et~al. 2016.
\newblock Abstractive text summarization using sequence-to-sequence rnns and
  beyond.
\newblock \emph{arXiv preprint arXiv:1602.06023}.

\bibitem[{Patro et~al.(2020)Patro, Kurmi, Kumar, and
  Namboodiri}]{patro2020deep}
Patro, B.; Kurmi, V.; Kumar, S.; and Namboodiri, V. 2020.
\newblock Deep bayesian network for visual question generation.
\newblock In \emph{Proceedings of the IEEE/CVF Winter Conference on
  Applications of Computer Vision}, 1566--1576.

\bibitem[{Ren et~al.(2015)Ren, He, Girshick, and Sun}]{ren2015faster}
Ren, S.; He, K.; Girshick, R.; and Sun, J. 2015.
\newblock Faster r-cnn: Towards real-time object detection with region proposal
  networks.
\newblock \emph{Advances in neural information processing systems}.

\bibitem[{Shazeer et~al.(2017)Shazeer, Mirhoseini, Maziarz, Davis, Le, Hinton,
  and Dean}]{shazeer2017outrageously}
Shazeer, N.; Mirhoseini, A.; Maziarz, K.; Davis, A.; Le, Q.; Hinton, G.; and
  Dean, J. 2017.
\newblock Outrageously large neural networks: The sparsely-gated
  mixture-of-experts layer.
\newblock \emph{arXiv preprint arXiv:1701.06538}.

\bibitem[{Singh et~al.(2019)Singh, Natarajan, Shah, Jiang, Chen, Batra, Parikh,
  and Rohrbach}]{singh2019towards}
Singh, A.; Natarajan, V.; Shah, M.; Jiang, Y.; Chen, X.; Batra, D.; Parikh, D.;
  and Rohrbach, M. 2019.
\newblock Towards vqa models that can read.
\newblock In \emph{Proceedings of the IEEE/CVF Conference on Computer Vision
  and Pattern Recognition}.

\bibitem[{Tan and Bansal(2019)}]{tan2019lxmert}
Tan, H.; and Bansal, M. 2019.
\newblock LXMERT: Learning Cross-Modality Encoder Representations from
  Transformers.
\newblock In \emph{Proceedings of the 2019 Conference on Empirical Methods in
  Natural Language Processing and the 9th International Joint Conference on
  Natural Language Processing}.

\bibitem[{Vaswani et~al.(2017)Vaswani, Shazeer, Parmar, Uszkoreit, Jones,
  Gomez, Kaiser, and Polosukhin}]{vaswani2017attention}
Vaswani, A.; Shazeer, N.; Parmar, N.; Uszkoreit, J.; Jones, L.; Gomez, A.~N.;
  Kaiser, {\L}.; and Polosukhin, I. 2017.
\newblock Attention is all you need.
\newblock \emph{Advances in neural information processing systems}, 30.

\bibitem[{Vinyals, Fortunato, and Jaitly(2015)}]{vinyals2015pointer}
Vinyals, O.; Fortunato, M.; and Jaitly, N. 2015.
\newblock Pointer networks.
\newblock \emph{Advances in neural information processing systems}.

\bibitem[{Wang et~al.(2021)Wang, Bao, Dong, and Wei}]{wang2021vlmo}
Wang, W.; Bao, H.; Dong, L.; and Wei, F. 2021.
\newblock VLMo: Unified Vision-Language Pre-Training with
  Mixture-of-Modality-Experts.
\newblock \emph{arXiv preprint arXiv:2111.02358}.

\bibitem[{Yu et~al.(2017)Yu, Fu, Mei, and Rui}]{yu2017multi}
Yu, D.; Fu, J.; Mei, T.; and Rui, Y. 2017.
\newblock Multi-level attention networks for visual question answering.
\newblock In \emph{IEEE conference on computer vision and pattern recognition}.

\bibitem[{Yu et~al.(2021)Yu, Tang, Yin, Sun, Tian, Wu, and Wang}]{yu2021ernie}
Yu, F.; Tang, J.; Yin, W.; Sun, Y.; Tian, H.; Wu, H.; and Wang, H. 2021.
\newblock Ernie-vil: Knowledge enhanced vision-language representations through
  scene graphs.
\newblock In \emph{Proceedings of the AAAI Conference on Artificial
  Intelligence}, 4.

\bibitem[{Zhang et~al.(2021)Zhang, Li, Hu, Yang, Zhang, Wang, Choi, and
  Gao}]{zhang2021vinvl}
Zhang, P.; Li, X.; Hu, X.; Yang, J.; Zhang, L.; Wang, L.; Choi, Y.; and Gao, J.
  2021.
\newblock Vinvl: Revisiting visual representations in vision-language models.
\newblock In \emph{IEEE/CVF Conference on Computer Vision and Pattern
  Recognition}.

\bibitem[{Zhu et~al.(2020)Zhu, Gao, Wang, and Wu}]{zhu2020simple}
Zhu, Q.; Gao, C.; Wang, P.; and Wu, Q. 2020.
\newblock Simple is not easy: A simple strong baseline for textvqa and
  textcaps.
\newblock \emph{arXiv preprint arXiv:2012.05153}, 2.

\end{thebibliography}

\end{document}